\pdfoutput=1

\documentclass[11pt]{article}

\usepackage{EMNLP2023}

\usepackage{times}
\usepackage{latexsym}

\usepackage[T1]{fontenc}

\usepackage[utf8]{inputenc}

\usepackage{microtype}

\usepackage{inconsolata}

\usepackage{multirow}
\usepackage{tabularx}
\usepackage{booktabs}
\usepackage{graphicx}

\usepackage{enumitem}
\usepackage{booktabs}
\usepackage{array}
\usepackage{hyperref}
\usepackage{url}
\usepackage{multirow}
\usepackage{colortbl}
\usepackage{booktabs}
\usepackage{amssymb}
\usepackage{amsmath}
\usepackage{amsthm}
\usepackage{graphicx}
\usepackage{makecell}
\usepackage{threeparttable}
\usepackage{CJKutf8}
\usepackage{lscape}
\usepackage{fancyhdr}
\usepackage[ruled,vlined]{algorithm2e}

\usepackage{threeparttable}

\usepackage{caption}
\usepackage{graphicx}
\usepackage{float} 
\usepackage{subfigure}
\usepackage{soul}

\usepackage{multicol}
\usepackage{adjustbox}
\usepackage{booktabs}

\title{InstructUIE: Multi-task Instruction Tuning for Unified Information Extraction}

\author{
    {\normalsize
     \textbf{Xiao Wang}$^{\bigstar*}$, 
     \ \ Weikang Zhou$^{\bigstar}$\thanks{$^*$  Equal contribution.} ,
     \ \ Can Zu$^{\bigstar}$, 
     \ \ Han Xia$^{\bigstar}$, 
     \ \ Tianze Chen$^{\bigstar}$,}\\
    {\normalsize
     \textbf{Yuansen Zhang}$^{\bigstar}$, 
    \ \ \textbf{Rui Zheng}$^{\bigstar}$, 
    \ \ \textbf{Junjie Ye}$^{\bigstar}$, 
    \ \ \textbf{Qi Zhang}$^{\bigstar}$$^{\dagger}$,
    \ \ \textbf{Tao Gui}$^{\blacklozenge}$
    \thanks{{} {} Corresponding Author} 
    \textbf{,}
    }\\
    {\normalsize
    \textbf{Jihua Kang}$^{\clubsuit}$\textbf{,} 
    \ \ \textbf{Jingsheng Yang}$^{\clubsuit}$\textbf{,} 
    \ \ \textbf{Siyuan Li}$^{\clubsuit}$\textbf{,} 
    \ \ \textbf{Chunsai Du}$^{\clubsuit}$\textbf{,}
    }\\
  {$^\bigstar$ \normalsize School of Computer Science, Fudan University, Shanghai, China} \\
  {$^\blacklozenge$ \normalsize Institute of Modern Languages and Linguistics, Fudan University, Shanghai, China} \\
  {$^\clubsuit$ \normalsize ByteDance Inc.} \\
  \texttt{\normalsize \{xiao\_wang20,qz,tgui\}@fudan.edu.cn}
}

\begin{document}
\maketitle
\begin{abstract}

Large language models have unlocked strong multi-task capabilities from reading instructive prompts.
However, recent studies have shown that existing large models still have difficulty with information extraction tasks. 
For example, gpt-3.5-turbo achieved an F1 score of 18.22 on the Ontonotes dataset, which is significantly lower than the state-of-the-art performance.
In this paper, we propose InstructUIE, a unified information extraction framework based on instruction tuning, which can uniformly model various information extraction tasks and capture the inter-task dependency.
To validate the proposed method, we introduce IE INSTRUCTIONS, a benchmark of 32 diverse information extraction datasets in a unified text-to-text format with expert-written instructions.
Experimental results demonstrate that our method achieves comparable performance to Bert in supervised settings and significantly outperforms the state-of-the-art and gpt3.5 in zero-shot settings.

\end{abstract}

\section{Introduction}

Large language models (LLMs) \cite{Brown2020LanguageMA,InstructGPT,GPT4} show tremendous promise in generalization within the set of observed tasks through multi-task training and unified encoding \cite{mishra-etal-2022-cross,wang-etal-2022-super,Longpre2023TheFC}. 
Recent research has revealed a significant performance gap in LLMs when it comes to information extraction (IE) tasks \cite{ye2023comprehensive,chen2023robust}. 
For instance, gpt-3.5-turbo achieves an 18.22 F1 score on the Ontonotes dataset, which is far from satisfactory. 
Therefore, it is necessary to explore how to build a unified information extraction (UIE) model with LLMs.

\begin{figure}[t]
\small
\centering
  \includegraphics[width=3.0in]{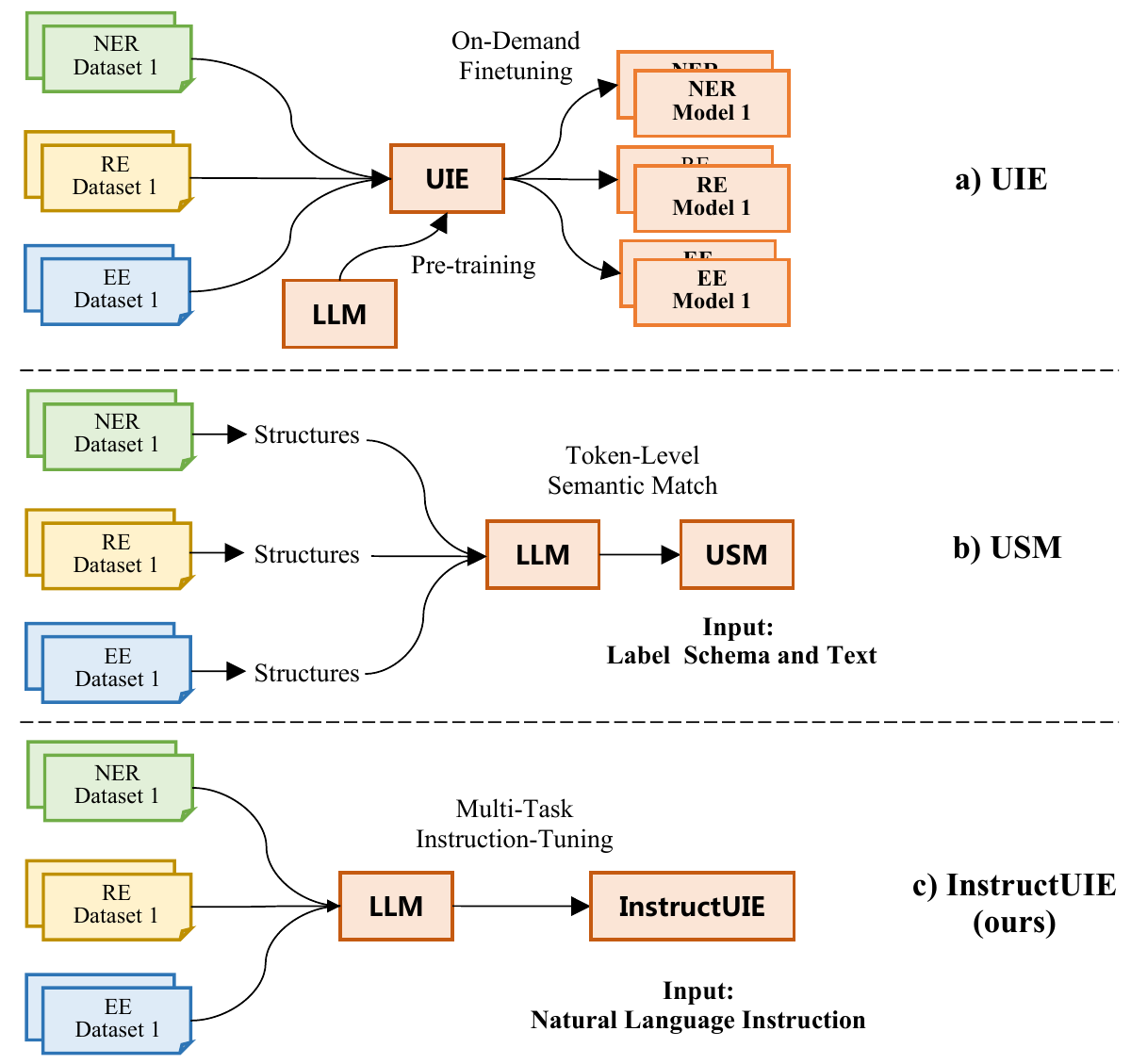}
  \caption{Illustration of 3 different paradigms for solving unified information extraction task.}
 \label{comparison}
\end{figure}

Recently, \citet{UIE} proposed UIE, which uniformly encodes different extraction structures via a structured extraction language, and captures the common IE abilities via a large-scale pretrained text-to-structure model (shown in Figure \ref{comparison}a). However, UIE requires separate finetune for different downstream tasks. This lead to the poor performance of UIE in low resource settings or facing new label schema, which greatly restricts the application of UIE in real scenarios. 
\citet{USM} proposed USM, which decouple IE into two basic tasks, token-token linking to extract label-agnostic substructures, and label-token linking to attach substructures to corresponding semantic concepts (shown in Figure \ref{comparison}b). However, USM presents two major limitations. Firstly, it converts IE into a semantic matching task, which makes it difficult to integrate with generative language model. Secondly, the method requires semantic matching for each word, which leads to a significant increase in training and inference time.

\begin{figure*}[t]
\centering
  \includegraphics[width=6.5in]{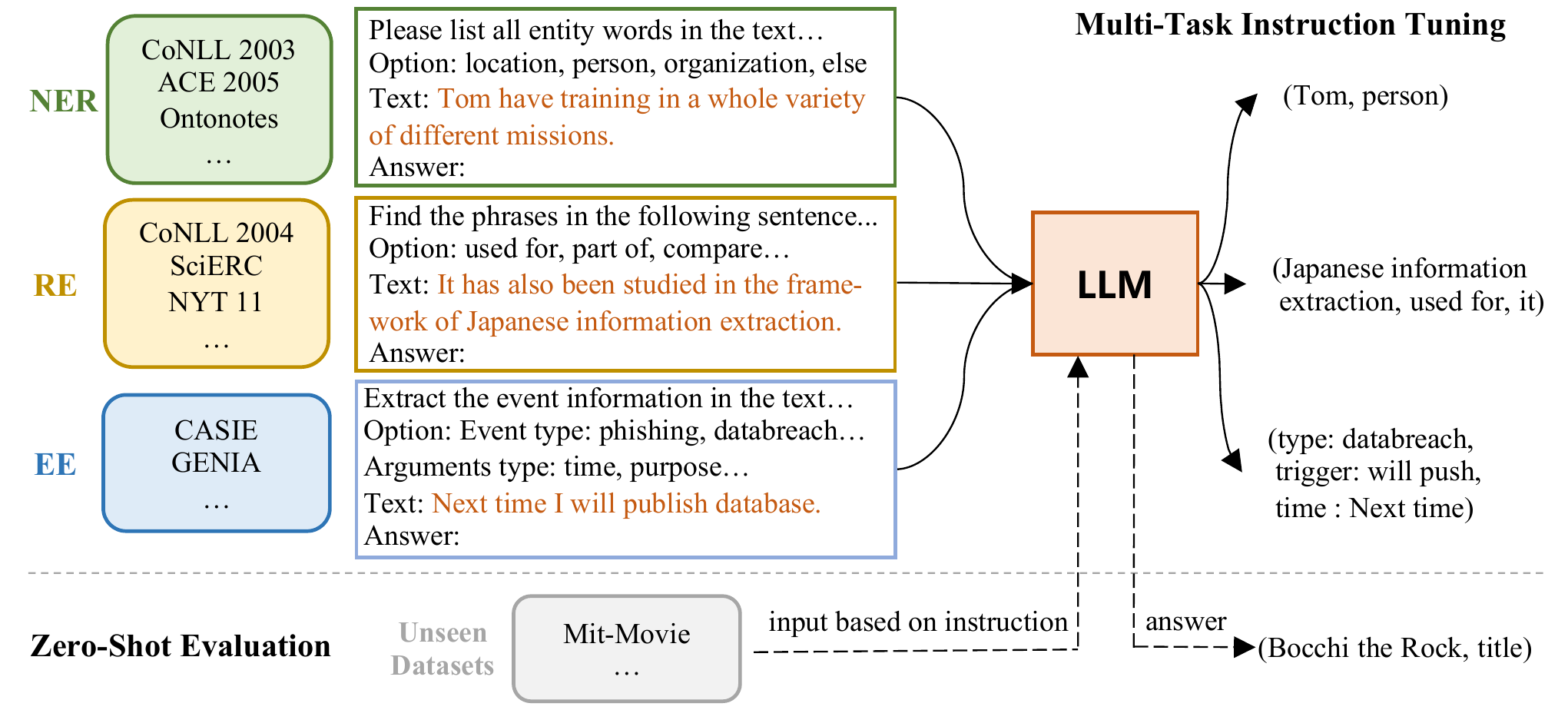}
  \caption{The overview framework of InstructUIE. The input consists of task instructions, options, and text. The output is a more understandable sentence converted from the original label structures.}
 \label{fig:framework}
\end{figure*}

In this work, we introduce a unified information extraction framework based on multi-task instruction tuning, named InstructUIE (shown in Figure \ref{comparison}c). Specifically, we reformulate IE tasks as a natural language generation problem. 
For the source sentence, we design descriptive instructions to enable the model to understand different tasks and employ an option mechanism including all candidate categories as constraints of output space. 
Then, a pre-trained language model is required to generate the target structure and the corresponding type in the form of natural language. 
We believe that unrestricted decoding would stimulate the latent knowledge of LLMs to complete IE tasks to a larger extent. 
We further propose auxiliary tasks, which enable the model to capture common structure information and deepen the understanding of diverse semantics. 
Specifically, we introduce entity span extraction task and entity typing task for named entity recognition (NER) task, entity pair extraction task and entity pair relationship identification task for relation extraction (RE) task, and trigger extraction task and argument extraction task for event extraction (EE) task.

To evaluate the effectiveness of the proposed model, we have developed a new benchmark called IE INSTRUCTIONS. The benchmark consists of 32 diverse information extraction datasets that have been unified into a text-to-text format, allowing for a consistent and standardized evaluation of various IE tasks \footnote{The dataset, code, and models can be found at 
https://github.com/BeyonderXX/InstructUIE}. 
Based on the benchmark, we conduct experiments on three main IE tasks under the supervised and zero-shot settings.

Our main contributions are summarized as follows:

\begin{itemize}[leftmargin=*, align=left]
    \item We propose an end-to-end framework for universal information extraction – InstructUIE, which leverages natural language instructions to guide large language models for IE tasks.
    \item We introduce IE INSTRUCTIONS, a benchmark of 32 diverse information extraction datasets in a unified text-to-text format with expert-written instructions. 
    \item Experimental results demonstrate that InstructUIE achieves comparable performance to Bert in a supervised setup. Notably, our method significantly outperforms the current state-of-the-art and GPT-3.5 in a zero-shot setup.
\end{itemize}

\section{Methodology}
In this section, we first briefly introduce the setup of instruction tuning. 
Then, we discuss the task meta-information schema and how IE tasks are mapped into our schema. 
Next, we discuss the framework of InstructUIE, which consists of two major parts: task schema and auxiliary tasks.
Finally, we explain how IE INSTRUCTION is constructed.

\subsection{Instruction Tuning Background}
Instruction tuning is a multi-task learning framework that enables the use of human-readable instructions to guide the output of LLMs. Given a source text and task-specific instructions, the model is trained to generate a sequence of tokens representing the desired output structure and its corresponding labels. 

In a supervised setup, the instructions are provided during training for all tasks, and the model is fine-tuned on a set of labeled data for each task. This allows the model to learn task-specific features and optimize for each task. In a zero-shot setup, the instructions are only provided for a subset of tasks during training, and the model is evaluated on unseen tasks without additional fine-tuning. This requires the model to generalize across tasks and use the shared features learned from the instruction tuning framework to infer the output structures for new tasks.

\subsection{Framework}
In this section, we discuss the task meta-information schema and how IE tasks are mapped into our schema. 
Next, we propose auxiliary tasks, which enable the model to capture common structure information and deepen the understanding of diverse semantics.

\subsubsection{Task Schema}\label{task schema}
To better transfer and utilize the knowledge learned in pre-trained language models, we reformulate the IE tasks to the seq2seq form and solve it through fine-tuning LLMs, as shown in Figure \ref{fig:framework}. Every task instance is formatted with four properties: task instruction, options, text, and output.

\textbf{Task Instruction} provides a detailed guide on how to extract the relevant information from the input text and produce the desired output structure. It includes information such as the type of information to be extracted, the format of the output structure, and any additional constraints or rules that need to be followed during the extraction process. The task instruction acts as a bridge between the raw input text and the structured output representation, enabling the model to understand the extraction task and generate accurate and meaningful output. In Table \ref{prompts_details} in the Appendix we present the list of instructions for each task.

\textbf{Options} are the output label constraints for a task, which represent the set of possible outputs that can be generated by the model for a given input. These label constraints are specific to each task and provide information on how to map the predicted outputs to the corresponding semantic concepts. For instance, in NER, options could be entity tags such as person, organization, location, or miscellaneous. Similarly, in RE, options could represent the types of relations that can be extracted, such as "works for", "born in", "married to", and so on. In EE, options could represent the event tags that correspond to different types of events, such as "beginning", "end", "occurring", "ceasing", and so on. The options provide a structured output space for the model, allowing it to generate outputs that are consistent with the underlying semantic structure of the task.

\textbf{Text} is the input sentence of a task instance. This sequence is then fed into the pre-trained language model along with the task instruction and options, enabling the model to generate the desired output sequence for the given task.

\textbf{Output} is the sentence converted from the original tags of the sample. Specifically, for NER, the output format is "\textit{entity tag: entity span}". For RE, the output format is "\textit{relationship: head entity, tail entity}". For EE, the output format is "\textit{event tag: trigger word, argument tag: argument span}". In cases where the input does not contain structural information that matches any of the provided options, we assign a value of "\textit{None}" to the corresponding output sentence.

\subsubsection{Auxiliary Tasks}
To boost the performance in a more fine-grained level, we further design auxiliary tasks to be trained in conjunction with the main task. The auxiliary tasks provide additional information that complements the main task, enabling the model to capture common structures better and deepen the understanding of diverse semantics.

For the named entity recognition task, we introduce a span extraction task and an entity typing task. The span extraction task is designed to extract the entity span from the input sentence, while the entity typing task is aimed at identifying the type of entity.

For the relation extraction task, we have introduced an entity pair extraction task and a relation classification task. The entity pair extraction task aims to extract the entity pairs involved in the relationship, while the relation classification task is designed to classify the type of relationship between the entity pairs.

For the event extraction task, we have introduced a trigger extraction task and an argument extraction task. The trigger extraction task is designed to extract the trigger word that triggers the event, while the argument extraction task aims to extract the associated arguments.

\subsection{IE INSTRUCTIONS}
\begin{figure}[t]
\small
\centering
  \includegraphics[width=3.0in]{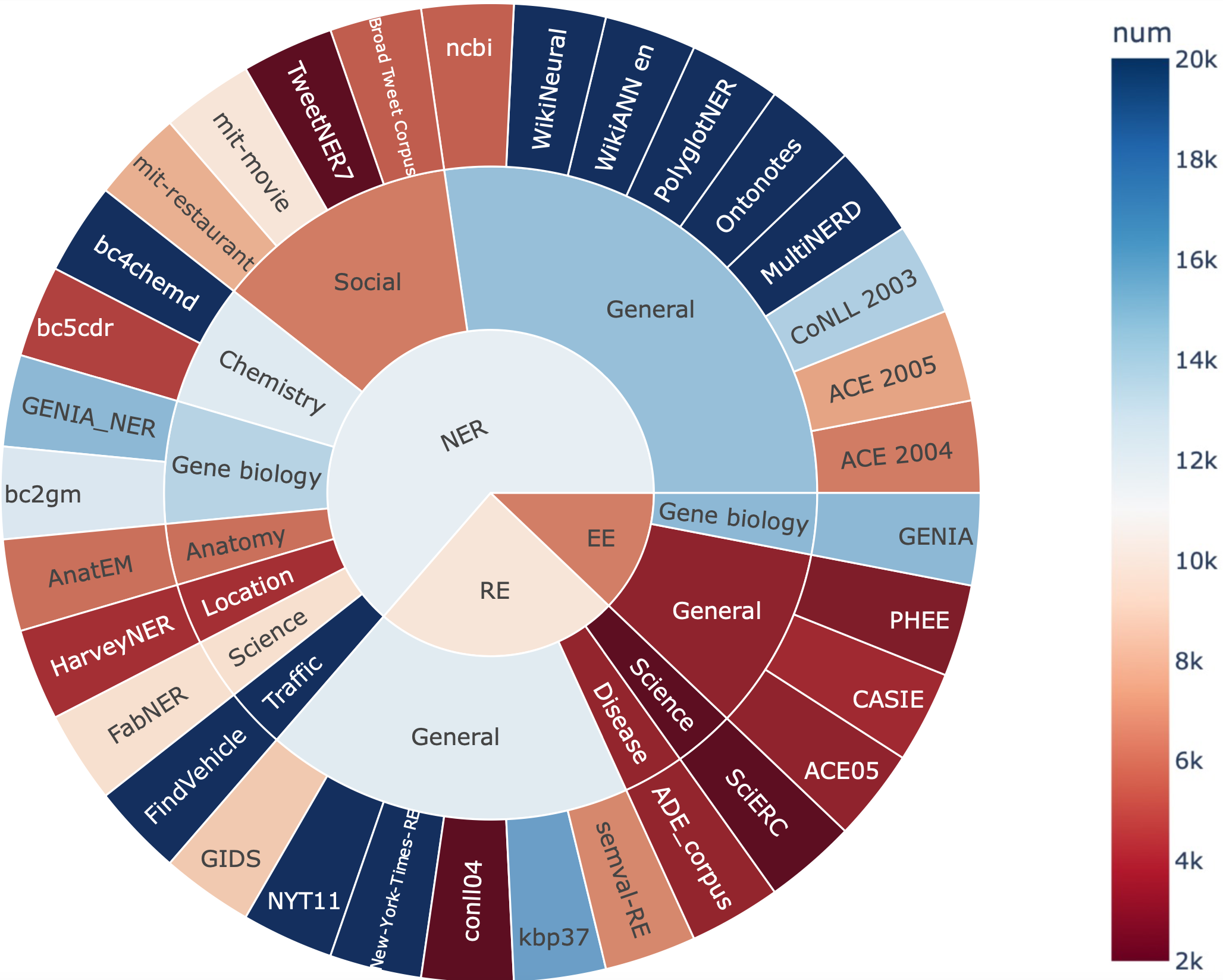}
  \caption{Overview of IE INSTRUCTIONS.}
 \label{benchmark_sun}
\end{figure}

IE INSTRUCTIONS collects 32 publicly available datasets covering three types of IE tasks: NER, RE, and EE. To ensure the diversity of the datasets, we include corpora from various domains, such as science, healthcare, social media, and transportation, in addition to general-domain sources, such as news and Wikidata. Figure \ref{benchmark_sun} shows the breakdown of the benchmark by task, domain, and size. For detailed dataset statistics and train/test split methods, please refer to Appendix Table \ref{dataset-details}.

We carry out the following data processing steps: (1) To address the issue of inconsistent label schemas across different tasks, we unify the names of labels with identical semantics but different names in various datasets. (2) To better test the semantic understanding capabilities of the LLM, we convert labels with underscores, abbreviations, or special formats into natural language formats. For example, we renamed the label "people person place\_of\_birth" to "place of birth." (3) Following the guidelines outlined in the section \ref{task schema}, we transform all datasets into a text-to-text format, which ensures a consistent representation of the input-output pairs across all tasks.

Our benchmark provides a standardized evaluation platform for LLMs' performance on IE tasks. This will facilitate a more accurate comparison of various models and contribute to the development of more effective and robust models for IE tasks.

\section{Experiments}
This section conducted extensive experiments under supervised and zero-shot settings to validate the effectiveness of InstructUIE. 
We select 11B FlanT5 \cite{chung2022scaling} as our backbone model because prior research \cite{Longpre2023TheFC} has demonstrated that models fine-tuned on instruction-based tasks offer a computationally efficient starting point for new tasks.
The details of the experimental setup, datasets, and comparison methods are described in the following parts.


\begin{table}[t]
    \centering
    \adjustbox{max width=\columnwidth}{
    \begin{tabular}{c|ccc|c}
    \toprule  
    Dataset & UIE & USM & Bert-base & Ours \\
    \midrule
    ACE2005 & 85.78 & 87.14 & \textbf{87.30} & 86.66 \\
    AnatEM & - & - & 85.82 & \textbf{90.89} \\
    bc2gm & - & - & 80.90 & \textbf{85.16} \\
    bc4chemd & - & - & 86.72 & \textbf{90.30} \\
    bc5cdr & - & - & 85.28 & \textbf{89.59} \\
    broad twitter & - & - & 58.61 & \textbf{83.14} \\
    CoNLL2003 & 92.99 & \textbf{93.16} & 92.40 & 92.94 \\
    FabNER & - & - & 64.20 & \textbf{76.20} \\
    FindVehicle & - & - & 87.13 & \textbf{89.47} \\
    GENIA-Ent & - & - & 73.3 & \textbf{74.71} \\
    HarveyNER & - & - & 82.26 & \textbf{88.79} \\
    MIT Movie & - & -& 88.78 & \textbf{89.01} \\
    MIT Restaurant & - & - & 81.02 & \textbf{82.55} \\
    multiNERD & - & - & 91.25 & \textbf{92.32} \\
    ncbi-disease & - & - & 80.20 & \textbf{90.23} \\
    Ontonotes & - & - & \textbf{91.11} & 90.19 \\
    polyglot-NER & - & - & \textbf{75.65} & 70.15 \\
    tweetNER7 & - & - & 56.49 & \textbf{64.97} \\
    wikiann & - & - & 70.60 & \textbf{85.13} \\
    wikineural & - & - & 82.78 & \textbf{91.36} \\
    Avg & - & - & 80.09 & \textbf{85.19} \\
    \bottomrule
    \end{tabular}
    }
    \caption{
Overall results of InstructUIE on NER task. The evaluation metric is Entity F1. For 20 NER datasets, InstructUIE outperforms the Bert model on 17 of them.
}
\label{supervised-result-NER}
\end{table}

\subsection{Experiments on Supervised Settings}
\subsubsection{Dataset}
We conduct supervised experiments on IE INSTRUCTIONS, including three tasks (named entity extraction, relation extraction, and event extraction). 
Details of the dataset splitting methods and statistics can be found in Appendix \ref{data details}.

To balance the dataset, we apply a sampling strategy \cite{poolsawad2014balancing}. Specifically, we sample 10,000 examples for each dataset and include all examples for datasets with fewer than 10,000 samples.

\subsubsection{Baselines}
\label{supervise-baseline}
We compare the proposed InstructUIE with the following strong baseline models:
\begin{itemize}
\item \textbf{UIE} \cite{UIE} is a unified text-to-structure generation framework that can universally model different IE tasks and adaptively generate targeted structures;
\item \textbf{USM} \cite{USM} is a unified IE tasks framework, which converts IE tasks to a semantic matching problem;
\item \textbf{Bert} \cite{devlin-etal-2019-bert}, which are widely used as text encoders for various tasks.
\end{itemize}

\subsubsection{Evaluation Metrics}
We use span-based offset Micro-F1 as the primary metric to evaluate the model. For NER task, we follow a span-level evaluation setting, where the entity boundary and entity type must be correctly predicted. For RE task, a relation triple is correct if the model correctly predicts the boundaries of the subject entity, the object entity, and the entity relation. For EE task, we report two evaluation metrics: (1) Event Trigger: an event trigger is correct if the event type and the trigger word are correctly predicted. (2) Event Argument: an event argument is correct if its role type and event type match a reference argument mention.

\subsubsection{Results}

\begin{table}[t]
    \centering
    \adjustbox{max width=\columnwidth}{
    \begin{tabular}{c|ccc}
    \toprule  
    Dataset & UIE & USM & Ours \\
    \midrule
    ADE corpus & - & -  & \textbf{82.31} \\
    CoNLL2004 & 75.00 & \textbf{78.84}  & 78.48 \\
    GIDS & - & -  & \textbf{81.98} \\
    kbp37 & - & - & \textbf{36.14} \\
    NYT & - & - & \textbf{90.47} \\
    NYT11 HRL & - & - & \textbf{56.06} \\
    SciERC & 36.53 & 37.36 & \textbf{45.15} \\
    semeval RE & - & - & \textbf{73.23} \\
    Avg & - & - & \textbf{67.98} \\
    \bottomrule
    \end{tabular}
    }
    \caption{Overall results of InstructUIE on RE task. The evaluation metric is Relation Strict F1. Our model reaches an average F1 of 67.98\% on the eight datasets of the RE task and is comparable to the baseline.}
\label{supervised-result-RE}
\end{table}

\begin{table}[t]
    \centering
    \begin{subtable}
    \centering
    \begin{tabular}{c|ccc|c}
    \toprule  
    Dataset & UIE & USM & Bert-base & Ours \\
    \midrule
    ACE2005 & 73.36 & 72.41 & 72.5 & \textbf{77.13} \\
    CASIE & 69.33 & \textbf{71.73} & 68.98 & 67.80 \\
    PHEE & - & - & - & \textbf{70.14} \\
    Avg & - & - & - & \textbf{71.69} \\
    \bottomrule
    \end{tabular}
    \caption*{a. Event Trigger F1}
    \label{subtable1}
    \end{subtable}
    \hspace{0.1cm}
    \begin{subtable}
    \centering
    \begin{tabular}{c|ccc|c}
    \toprule  
    Dataset & UIE & USM & Bert-base & Ours \\
    \midrule
    ACE2005 & 54.79 & 55.83 & 59.9 & \textbf{72.94} \\
    CASIE & 61.30 & 63.26 & 60.37 & \textbf{63.53} \\
    PHEE & - & - & - & \textbf{62.91} \\
    Avg & - & - & - & \textbf{66.46} \\
    \bottomrule
    \end{tabular}
    \caption*{b. Event Argument F1}
    \label{subtable2}
    \end{subtable}
    \caption{Overall results of InstructUIE on EE task. The evaluation metric is Event Trigger F1 and Event Argument F1. Our model outperformed USM and UIE on some datasets.}
    \label{supervised-result-EE}
\end{table}


Tabel \ref{supervised-result-NER}, Tabel \ref{supervised-result-RE} and \ref{supervised-result-EE} show the performance of different models for the NER, RE, and EE tasks. 
\paragraph{Named Entity Recognition} 
Our model achieves an average F1 score of 85.19\% on 20 NER datasets, surpassing Bert's 80.09\%. The best performance is on the CoNLL2003 dataset, where InstructUIE achieved an F1 score of 92.94\%. For 20 NER data sets, InstructUIE outperforms the Bert model on 17 of them. Among them, our model outperforms Bert by more than 5 points on eight datasets. The dataset with the biggest gap is the broad twitter dataset, where InstructUIE outperforms Bert by about 25 points.

In the ACE2005, Ontonotes, and Polyglot-NER datasets, our model performs slightly worse than Bert. We speculate that this is due to our strategy of sampling only 10,000 training examples for each dataset. The original corpora for these three datasets contain a larger number of training examples, such as 420,000 for Polyglot-NER, of which we only used around 20\%. The detailed number of training sets for all datasets can be seen in the appendix.

Compared with UIE and USM, our model also achieves comparable results on ACE2005 and CoNLL2003, which are two commonly used datasets. Due to the UIE and USM only test their modelson a small number of commonly used datasets, we are unable to compare our model with these two models on other datasets.

\begin{table*}[htbp]
    \centering
    \begin{tabular}{c|ccccccc}
    \toprule
        Model & Movie & Restaurant & AI & Literature & Music & Politics & Science \\
        \midrule
        USM & 37.73 & 14.73 & 28.18 & \textbf{56.00} & 44.93 & 36.10 & 44.09 \\
        InstructUIE & \textbf{63.00} & \textbf{20.99} & \textbf{49.00} & 47.21 & \textbf{53.16} & \textbf{48.15} & \textbf{49.30} \\
    \bottomrule
    \end{tabular}
    \caption{
    Micro-F1 scores of zero-shot NER on 7 datasets. The best results are in bold. InstructUIE outperforms SOTA by a wide margin on most datasets ranging from 5.21\% to 25.27\%.}
    \label{zero-shot-NER}
\end{table*}

\begin{table*}[t]
    \centering
    \begin{tabular}{c|ccc}
    \toprule
         & Model & FewRel & Wiki-ZSL \\
        \midrule
        \multirow{2}*{Baselines} & $ZETT_{T5-small}$ & 30.53 & 31.74 \\
        ~ & $ZETT_{T5-base}$ & 33.71 & 31.17 \\
        \midrule
        Ours & InstructUIE & \textbf{39.55} & \textbf{ 35.20} \\
    \bottomrule
    \end{tabular}
    \caption{
    Micro-F1 scores of zero-shot RE on FewRel and Wiki-ZSL. The best results are in bold. InstructUIE outperforms SOTA on both datasets.}
    \label{zero-shot-RE}
\end{table*}

\paragraph{Relational Extraction}
Our model reaches an average F1 of 67.98\% on the eight datasets of the RE task, among which the NYT data set reaches 90.47\% F1 score. Among the eight datasets, CoNLL2004 and SciERC datasets are also tested by UIE and USM models. We focus on the analysis of the results of these two datasets. For the SciERC dataset, InstructUIE significantly outperforms UIE and USM by 8.62\% and 7.79\% respectively. For the CoNLL2004 dataset, InstructUIE outperforms UIE by more than three points, and lag USM by less than 0.5\%.
Moreover, noted that as BERT is usually used for relation classification tasks rather than relation extraction. Therefore, we did not use this baseline in the RE task.


\paragraph{Event Extraction}
Our model achieve sota on all datasets except for the Event Trigger F1 metric of the CASIE dataset. On the Event Trigger F1 metric, InstructUIE reaches an average of 71.69\% on these three datasets, with ACE2005 reaching 77.13\%, significantly surpassing UIE's 73.36\%,  USM's 72.41\% and Bert's 72.5\%. On the Event Argument F1 metric, InstructUIE beats three baseline models to reach sota on all three datasets. In particular, ACE2005 dataset reaches 72.94\%, 18 points higher than the UIE and 17 points higher than the USM.

\subsection{Experiments on Zero-shot Settings}

\subsubsection{Dataset}
To evaluate InstructUIE's zero-shot performance, we train the model on 18 NER datasets and 6 RE datasets and test it on 7 NER datasets and 2 RE datasets. Specifically, we eliminate the datasets for zero-shot experimental testing during the training phase. 
For the NER task, We use five CrossNER subsets(AI, literature, music, politics, science) \cite{CrossNERDATASET}, MIT Movie Review, and MIT Restaurant Review \cite{MITReviewDataset} to test the zero-shot capability of the model. 
For RE task, we test the zero-shot capability on FewRel \cite{FewRelDATASET} and Wiki-ZSL \cite{Wiki-ZSLDATASET}. For FewRel and Wiki-ZSL data sets, we follow the previous work \cite{relationprompt} and randomly select 5 unseen labels which do not appear in the training set as the test set. In order to reduce the effect of experimental noise, the unseen label selection process is repeated for five different random seeds to produce the test set.

Since the training and testing tasks do not overlap at all and across various domains as well, this setting is challenging.

\subsubsection{Baselines}
For zero-shot Named Entity Recognition and Relational Extraction, we compare InstructUIE with the following strong baselines:
\begin{itemize}
\item \textbf{ZETT}\cite{zett} is a novel framework based on end-to-end generative transformers and outperform previous state-of-the-art models;
\item \textbf{ChatGPT} \cite{InstructGPT} is also called GPT-3.5-turbo, which is the most capable GPT-3.5 model and optimized for chat;
\item \textbf{UIE} and \textbf{USM} have been introduced in \ref{supervise-baseline}.
\end{itemize}

\subsubsection{Results}

\begin{table*}[htbp]
    \centering
    \begin{tabular}{c|ccccccc|cc}
    \toprule
        Model & Movie & Restaurant & AI & Literature & Music & Politics & Science & FewRel & Wiki-ZSL \\
        \midrule
        davinci & 0.84 & 2.94 & 2.97 & 9.87 & 13.83 & 18.42 & 10.04 & 0.00 & 0.00 \\
        chatgpt & \textbf{41.00} & \textbf{37.76} & \textbf{54.40} & \textbf{54.07} & \textbf{61.24} & \textbf{59.12} & \textbf{63.00} & \textbf{9.96} & \textbf{13.14} \\
    \bottomrule
    \end{tabular}
    \caption{\label{zero-shot-GPT}
    Micro-F1 scores of davinci and chatgpt under zero-shot setting.}
\end{table*}

Table \ref{zero-shot-NER} and Table \ref{zero-shot-RE} show the performance of NER and RE tasks under the zero-shot setting. For the NER task, we can observe that InstructUIE outperforms the current sota model USM in Micro-F1 score on all the datasets except CrossNER\_Literature, ranging from 5.21\% to 25.27\%. For example, compared with the USM model, InstructUIE performs over 20 points better on the MIT Movie Review dataset and the CrossNER\_AI dataset. Noted that USM is trained on the same task corpus and tested on the label held out, while our model has never seen the task corpus. For the RE task, under the setting of 5 unseen labels, InstructUIE outperforms the current sota model ZETT on both the FewRel and Wiki-ZSL datasets by 5.84\% and 3.46\% respectively.

When compared to the GPT series model, InstructUIE significantly outperforms Davinci for the NER task but still falls some way short of Chatgpt's results for the NER task. However, for the RE task, our model performs much better than these two GPT series models. Both Davinci and Chatgpt perform poorly, especially with Davinci completely unable to output correct results.

It is worth mentioning that since Chatgpt is not open source, we have no way of knowing whether the model has seen the two data sets used by the zero-shot setting during training, and we think the huge difference in results for NER and RE tasks may be due to this reason.

\section{Related Work}

\subsection{Instruction Tuning}
Instruction tuning \cite{mishra-etal-2022-cross, wang-etal-2022-super, Longpre2023TheFC}, a novel paradigm that leverages natural language instructions to guide large language models for downstream tasks, shows tremendous promise in generalization within the set of observed tasks.
Most recent work \cite{wang-etal-2022-super,Longpre2023TheFC} on instruction tuning has focused on general NLP tasks such as question answering and text classification, but not specifically on IE tasks.
While some work such as \cite{Wang2022InstructionNERAM, parmar-etal-2022-boxbart} includes a few IE tasks, those tasks do not provide good coverage of IE tasks and domains. 
No prior work has examined how training a model on a wide range of IE tasks with various instructions.
In this paper, we propose a unified framework for information extraction that involves auxiliary task design as well as specific tuning methods.

\subsection{Information Extraction}
Information extraction is fundamental in natural language processing systems, aiming to extract structured information from unstructured or semi-structured data sources automatically. 
Traditional methods \cite{wang-etal-2022-miner,yan-etal-2021-unified,CoNLL2004SOTA,NYT11HRLSOTA} for IE typically require the design of specific architectures for different IE tasks, and the models are trained separately. However, training dedicated models for different IE tasks requires a significant amount of labeled data, which can be costly and time-consuming to obtain. Secondly, knowledge learned from one IE task cannot be easily applied to another task, even if the tasks have similar characteristics. 
Recently, \citet{UIE} proposed UIE, which uniformly encodes different extraction structures via a structured extraction language and captures the common IE abilities via a large-scale pre-trained text-to-structure model. However, UIE requires separate finetune for different downstream tasks. This lead to the poor performance of UIE in low resource settings or facing new label schema. 
\citet{USM} proposed USM, which decouples IE into two basic tasks, token-token linking and label-token linking.  Unfortunately, USM requires semantic matching for each word, which leads to a significant increase in training and inference time.
InstructUIE addresses these challenges by utilizing instructive guidance to direct pre-trained large models toward the task, facilitating the efficient and adaptive generation of target structures.

\section{Conclusion}
In this paper, we propose an end-to-end framework for universal information extraction – InstructUIE, which leverages natural language instructions to guide large language models for IE tasks. 
We further introduce a new benchmark dataset. The benchmark consists of 32 diverse information extraction datasets that have been unified into a text-to-text format, allowing for a consistent and standardized evaluation of various IE tasks.
Experimental results demonstrate that InstructUIE achieves state-of-the-art results under supervised and zero settings and solves massive tasks using a single multi-task model.


\bibliography{anthology,custom}

\begin{thebibliography}{75}
\expandafter\ifx\csname natexlab\endcsname\relax\def\natexlab#1{#1}\fi

\bibitem[{Al{-}Rfou et~al.(2014)Al{-}Rfou, Kulkarni, Perozzi, and
  Skiena}]{polyglot-NERDATASET}
Rami Al{-}Rfou, Vivek Kulkarni, Bryan Perozzi, and Steven Skiena. 2014.
\newblock \href {http://arxiv.org/abs/1410.3791} {{POLYGLOT-NER:} massive
  multilingual named entity recognition}.
\newblock \emph{CoRR}, abs/1410.3791.

\bibitem[{Brown et~al.(2020)Brown, Mann, Ryder, Subbiah, Kaplan, Dhariwal,
  Neelakantan, Shyam, Sastry, Askell, Agarwal, Herbert-Voss, Krueger, Henighan,
  Child, Ramesh, Ziegler, Wu, Winter, Hesse, Chen, Sigler, Litwin, Gray, Chess,
  Clark, Berner, McCandlish, Radford, Sutskever, and
  Amodei}]{Brown2020LanguageMA}
Tom~B. Brown, Benjamin Mann, Nick Ryder, Melanie Subbiah, Jared Kaplan,
  Prafulla Dhariwal, Arvind Neelakantan, Pranav Shyam, Girish Sastry, Amanda
  Askell, Sandhini Agarwal, Ariel Herbert-Voss, Gretchen Krueger, T.~J.
  Henighan, Rewon Child, Aditya Ramesh, Daniel~M. Ziegler, Jeff Wu, Clemens
  Winter, Christopher Hesse, Mark Chen, Eric Sigler, Mateusz Litwin, Scott
  Gray, Benjamin Chess, Jack Clark, Christopher Berner, Sam McCandlish, Alec
  Radford, Ilya Sutskever, and Dario Amodei. 2020.
\newblock Language models are few-shot learners.
\newblock \emph{ArXiv}, abs/2005.14165.

\bibitem[{Chen and Li(2021)}]{Wiki-ZSLDATASET}
Chih-Yao Chen and Cheng-Te Li. 2021.
\newblock \href {http://arxiv.org/abs/2104.04697} {Zs-bert: Towards zero-shot
  relation extraction with attribute representation learning}.

\bibitem[{Chen et~al.(2022{\natexlab{a}})Chen, Xu, Zhang, and
  Huang}]{HarveyNERDATASET}
Pei Chen, Haotian Xu, Cheng Zhang, and Ruihong Huang. 2022{\natexlab{a}}.
\newblock \href {https://doi.org/10.18653/v1/2022.naacl-main.243} {Crossroads,
  buildings and neighborhoods: A dataset for fine-grained location
  recognition}.
\newblock In \emph{Proceedings of the 2022 Conference of the North American
  Chapter of the Association for Computational Linguistics: Human Language
  Technologies}, pages 3329--3339, Seattle, United States. Association for
  Computational Linguistics.

\bibitem[{Chen et~al.(2022{\natexlab{b}})Chen, Xu, Zhang, and
  Huang}]{HarveyNERSOTA}
Pei Chen, Haotian Xu, Cheng Zhang, and Ruihong Huang. 2022{\natexlab{b}}.
\newblock \href {https://doi.org/10.18653/v1/2022.naacl-main.243} {Crossroads,
  buildings and neighborhoods: A dataset for fine-grained location
  recognition}.
\newblock In \emph{Proceedings of the 2022 Conference of the North American
  Chapter of the Association for Computational Linguistics: Human Language
  Technologies}, pages 3329--3339, Seattle, United States. Association for
  Computational Linguistics.

\bibitem[{Chen et~al.(2023)Chen, Ye, Zu, Xu, Zheng, Peng, Zhou, Gui, Zhang, and
  Huang}]{chen2023robust}
Xuanting Chen, Junjie Ye, Can Zu, Nuo Xu, Rui Zheng, Minlong Peng, Jie Zhou,
  Tao Gui, Qi~Zhang, and Xuanjing Huang. 2023.
\newblock How robust is gpt-3.5 to predecessors? a comprehensive study on
  language understanding tasks.
\newblock \emph{arXiv preprint arXiv:2303.00293}.

\bibitem[{Chia et~al.(2022)Chia, Bing, Poria, and Si}]{relationprompt}
Yew~Ken Chia, Lidong Bing, Soujanya Poria, and Luo Si. 2022.
\newblock \href {http://arxiv.org/abs/2203.09101} {Relationprompt: Leveraging
  prompts to generate synthetic data for zero-shot relation triplet
  extraction}.

\bibitem[{Chung et~al.(2022)Chung, Hou, Longpre, Zoph, Tay, Fedus, Li, Wang,
  Dehghani, Brahma et~al.}]{chung2022scaling}
Hyung~Won Chung, Le~Hou, Shayne Longpre, Barret Zoph, Yi~Tay, William Fedus,
  Eric Li, Xuezhi Wang, Mostafa Dehghani, Siddhartha Brahma, et~al. 2022.
\newblock Scaling instruction-finetuned language models.
\newblock \emph{arXiv preprint arXiv:2210.11416}.

\bibitem[{Derczynski et~al.(2016)Derczynski, Bontcheva, and
  Roberts}]{broad_twitter_corpusDATASET}
Leon Derczynski, Kalina Bontcheva, and Ian Roberts. 2016.
\newblock \href {https://aclanthology.org/C16-1111} {Broad {T}witter corpus: A
  diverse named entity recognition resource}.
\newblock In \emph{Proceedings of {COLING} 2016, the 26th International
  Conference on Computational Linguistics: Technical Papers}, pages 1169--1179,
  Osaka, Japan. The COLING 2016 Organizing Committee.

\bibitem[{Devlin et~al.(2019)Devlin, Chang, Lee, and
  Toutanova}]{devlin-etal-2019-bert}
Jacob Devlin, Ming-Wei Chang, Kenton Lee, and Kristina Toutanova. 2019.
\newblock \href {https://doi.org/10.18653/v1/N19-1423} {{BERT}: Pre-training of
  deep bidirectional transformers for language understanding}.
\newblock In \emph{Proceedings of the 2019 Conference of the North {A}merican
  Chapter of the Association for Computational Linguistics: Human Language
  Technologies, Volume 1 (Long and Short Papers)}, pages 4171--4186,
  Minneapolis, Minnesota. Association for Computational Linguistics.

\bibitem[{Dogan et~al.(2014)Dogan, Leaman, and Lu}]{ncbi-diseaseDATASET}
Rezarta~Islamaj Dogan, Robert Leaman, and Zhiyong Lu. 2014.
\newblock Ncbi disease corpus: A resource for disease name recognition and
  concept normalization.
\newblock \emph{Journal of biomedical informatics}, 47:1--10.

\bibitem[{Guan(2022)}]{FindVehicle}
Runwei Guan. 2022.
\newblock \href {github.com/GuanRunwei/FindVehicle} {Findvehicle and
  vehiclefinder: A ner dataset for a text-image cross-modal vehicle retrieval
  system}.

\bibitem[{Gurulingappa et~al.(2012)Gurulingappa, Rajput, Roberts, Fluck,
  Hofmann-Apitius, and Toldo}]{ADEcorpusDATASET}
Harsha Gurulingappa, Abdul~Mateen Rajput, Angus Roberts, Juliane Fluck, Martin
  Hofmann-Apitius, and Luca Toldo. 2012.
\newblock \href {https://doi.org/https://doi.org/10.1016/j.jbi.2012.04.008}
  {Development of a benchmark corpus to support the automatic extraction of
  drug-related adverse effects from medical case reports}.
\newblock \emph{Journal of Biomedical Informatics}, 45(5):885--892.
\newblock Text Mining and Natural Language Processing in Pharmacogenomics.

\bibitem[{Han et~al.(2018)Han, Zhu, Yu, Wang, Yao, Liu, and
  Sun}]{FewRelDATASET}
Xu~Han, Hao Zhu, Pengfei Yu, Ziyun Wang, Yuan Yao, Zhiyuan Liu, and Maosong
  Sun. 2018.
\newblock \href {http://arxiv.org/abs/1810.10147} {Fewrel: A large-scale
  supervised few-shot relation classification dataset with state-of-the-art
  evaluation}.

\bibitem[{Hendrickx et~al.(2010)Hendrickx, Kim, Kozareva, Nakov, S{\'e}aghdha,
  Pad{\'o}, Pennacchiotti, Romano, and Szpakowicz}]{Hendrickx2010SemEval2010T8}
Iris Hendrickx, Su~Nam Kim, Zornitsa Kozareva, Preslav Nakov, Diarmuid~{\'O}
  S{\'e}aghdha, Sebastian Pad{\'o}, Marco Pennacchiotti, Lorenza Romano, and
  Stan Szpakowicz. 2010.
\newblock Semeval-2010 task 8: Multi-way classification of semantic relations
  between pairs of nominals.
\newblock In \emph{*SEMEVAL}.

\bibitem[{Hovy et~al.(2006)Hovy, Marcus, Palmer, Ramshaw, and
  Weischedel}]{OntoNotesDataset}
Eduard~H. Hovy, Mitchell~P. Marcus, Martha Palmer, Lance~A. Ramshaw, and
  Ralph~M. Weischedel. 2006.
\newblock Ontonotes: The 90\% solution.
\newblock In \emph{North American Chapter of the Association for Computational
  Linguistics}.

\bibitem[{Huguet~Cabot and Navigli(2021)}]{CoNLL2004SOTA}
Pere-Llu{\'\i}s Huguet~Cabot and Roberto Navigli. 2021.
\newblock \href {https://doi.org/10.18653/v1/2021.findings-emnlp.204} {{REBEL}:
  Relation extraction by end-to-end language generation}.
\newblock In \emph{Findings of the Association for Computational Linguistics:
  EMNLP 2021}, pages 2370--2381, Punta Cana, Dominican Republic. Association
  for Computational Linguistics.

\bibitem[{Jat et~al.(2018)Jat, Khandelwal, and Talukdar}]{Jat2018ImprovingDS}
Sharmistha Jat, Siddhesh Khandelwal, and Partha~Pratim Talukdar. 2018.
\newblock Improving distantly supervised relation extraction using word and
  entity based attention.
\newblock \emph{ArXiv}, abs/1804.06987.

\bibitem[{Kim et~al.(2022)Kim, Iso, Bhutani, Hruschka, and Nakashole}]{zett}
Bosung Kim, Hayate Iso, Nikita Bhutani, Estevam Hruschka, and Ndapa Nakashole.
  2022.
\newblock \href {http://arxiv.org/abs/2212.10708} {Zero-shot triplet extraction
  by template infilling}.

\bibitem[{Kim et~al.(2003{\natexlab{a}})Kim, Ohta, Tateisi, and
  Tsujii}]{Kim2003GENIAC}
Jin-Dong Kim, Tomoko Ohta, Yuka Tateisi, and Junichi Tsujii.
  2003{\natexlab{a}}.
\newblock Genia corpus - a semantically annotated corpus for bio-textmining.
\newblock \emph{Bioinformatics}, 19 Suppl 1:i180--2.

\bibitem[{Kim et~al.(2003{\natexlab{b}})Kim, Ohta, Tateisi, and
  Tsujii}]{GENIANERDATASET}
Jin-Dong Kim, Tomoko Ohta, Yuka Tateisi, and Jun'ichi Tsujii.
  2003{\natexlab{b}}.
\newblock \href {https://doi.org/10.1093/bioinformatics/btg1023} {Genia
  corpus—a semantically annotated corpus for bio-textmining}.
\newblock \emph{Bioinformatics (Oxford, England)}, 19 Suppl 1:i180--2.

\bibitem[{Kocaman and Talby(2020{\natexlab{a}})}]{Kocaman2020BiomedicalNE}
Veysel Kocaman and David Talby. 2020{\natexlab{a}}.
\newblock Biomedical named entity recognition at scale.
\newblock In \emph{ICPR Workshops}.

\bibitem[{Kocaman and Talby(2020{\natexlab{b}})}]{bc2gmSOTA}
Veysel Kocaman and David Talby. 2020{\natexlab{b}}.
\newblock \href {http://arxiv.org/abs/2011.06315} {Biomedical named entity
  recognition at scale}.
\newblock \emph{CoRR}, abs/2011.06315.

\bibitem[{Kocaman and Talby(2020{\natexlab{c}})}]{ncbiSOTA}
Veysel Kocaman and David Talby. 2020{\natexlab{c}}.
\newblock \href {http://arxiv.org/abs/2011.06315} {Biomedical named entity
  recognition at scale}.
\newblock \emph{CoRR}, abs/2011.06315.

\bibitem[{Kocaman and Talby(2022{\natexlab{a}})}]{AnatEMSOTA}
Veysel Kocaman and David Talby. 2022{\natexlab{a}}.
\newblock \href {https://doi.org/https://doi.org/10.1016/j.simpa.2022.100373}
  {Accurate clinical and biomedical named entity recognition at scale}.
\newblock \emph{Software Impacts}, 13:100373.

\bibitem[{Kocaman and Talby(2022{\natexlab{b}})}]{bc4chemdSOTA}
Veysel Kocaman and David Talby. 2022{\natexlab{b}}.
\newblock \href {https://doi.org/https://doi.org/10.1016/j.simpa.2022.100373}
  {Accurate clinical and biomedical named entity recognition at scale}.
\newblock \emph{Software Impacts}, 13:100373.

\bibitem[{Krallinger et~al.(2015)Krallinger, Rabal, Leitner, Vazquez, Salgado,
  lu, Leaman, Lu, Ji, Lowe, Sayle, Batista-Navarro, Rak, Huber, Rocktäschel,
  Matos, Campos, Tang, Qi, and Valencia}]{bc4chemdDATASET}
Martin Krallinger, Obdulia Rabal, Florian Leitner, Miguel Vazquez, David
  Salgado, Zhiyong lu, Robert Leaman, Yanan Lu, Donghong Ji, Daniel Lowe, Roger
  Sayle, Riza Batista-Navarro, Rafal Rak, Torsten Huber, Tim Rocktäschel,
  Sérgio Matos, David Campos, Buzhou Tang, Wang Qi, and Alfonso Valencia.
  2015.
\newblock \href {https://doi.org/10.1186/1758-2946-7-S1-S2} {The chemdner
  corpus of chemicals and drugs and its annotation principles}.
\newblock \emph{Journal of Cheminformatics}, 7:S2.

\bibitem[{Kumar and Starly(2021{\natexlab{a}})}]{Kumar2021FabNERIE}
Aman Kumar and Binil Starly. 2021{\natexlab{a}}.
\newblock “fabner”: information extraction from manufacturing process
  science domain literature using named entity recognition.
\newblock \emph{Journal of Intelligent Manufacturing}, 33:2393 -- 2407.

\bibitem[{Kumar and Starly(2021{\natexlab{b}})}]{FabNERSOTA}
Aman Kumar and Binil Starly. 2021{\natexlab{b}}.
\newblock \href {https://doi.org/10.1007/s10845-021-01807-x} {“fabner”:
  information extraction from manufacturing process science domain literature
  using named entity recognition}.
\newblock \emph{Journal of Intelligent Manufacturing}, 33.

\bibitem[{Lai et~al.(2018)Lai, Leung, and Leung}]{semevalRESOTA}
Sunny Lai, Kwong~Sak Leung, and Yee Leung. 2018.
\newblock \href {https://doi.org/10.18653/v1/S18-1118} {{SUNNYNLP} at
  {S}em{E}val-2018 task 10: A support-vector-machine-based method for detecting
  semantic difference using taxonomy and word embedding features}.
\newblock In \emph{Proceedings of the 12th International Workshop on Semantic
  Evaluation}, pages 741--746, New Orleans, Louisiana. Association for
  Computational Linguistics.

\bibitem[{Li et~al.(2016)Li, Sun, Johnson, Sciaky, Wei, Leaman, Davis,
  Mattingly, Wiegers, and Lu}]{Li2016BioCreativeVC}
Jiao Li, Yueping Sun, Robin~J. Johnson, Daniela Sciaky, Chih-Hsuan Wei, Robert
  Leaman, Allan~Peter Davis, Carolyn~J. Mattingly, Thomas~C. Wiegers, and
  Zhiyong Lu. 2016.
\newblock Biocreative v cdr task corpus: a resource for chemical disease
  relation extraction.
\newblock \emph{Database: The Journal of Biological Databases and Curation},
  2016.

\bibitem[{Li et~al.(2019)Li, Sun, Meng, Liang, Wu, and Li}]{OntonotesSOTA}
Xiaoya Li, Xiaofei Sun, Yuxian Meng, Junjun Liang, Fei Wu, and Jiwei Li. 2019.
\newblock \href {http://arxiv.org/abs/1911.02855} {Dice loss for
  data-imbalanced {NLP} tasks}.
\newblock \emph{CoRR}, abs/1911.02855.

\bibitem[{Liu et~al.(2019)Liu, Meng, Zhang, Xu, Chen, and
  Zhou}]{MITReviewDataset}
Yijin Liu, Fandong Meng, Jinchao Zhang, Jinan Xu, Yufeng Chen, and Jie Zhou.
  2019.
\newblock \href {http://arxiv.org/abs/1906.02437} {{GCDT:} {A} global context
  enhanced deep transition architecture for sequence labeling}.
\newblock \emph{CoRR}, abs/1906.02437.

\bibitem[{Liu et~al.(2020)Liu, Xu, Yu, Dai, Ji, Cahyawijaya, Madotto, and
  Fung}]{CrossNERDATASET}
Zihan Liu, Yan Xu, Tiezheng Yu, Wenliang Dai, Ziwei Ji, Samuel Cahyawijaya,
  Andrea Madotto, and Pascale Fung. 2020.
\newblock \href {http://arxiv.org/abs/2012.04373} {Crossner: Evaluating
  cross-domain named entity recognition}.

\bibitem[{Longpre et~al.(2023)Longpre, Hou, Vu, Webson, Chung, Tay, Zhou, Le,
  Zoph, Wei, and Roberts}]{Longpre2023TheFC}
S.~Longpre, Le~Hou, Tu~Vu, Albert Webson, Hyung~Won Chung, Yi~Tay, Denny Zhou,
  Quoc~V. Le, Barret Zoph, Jason Wei, and Adam Roberts. 2023.
\newblock The flan collection: Designing data and methods for effective
  instruction tuning.
\newblock \emph{ArXiv}, abs/2301.13688.

\bibitem[{Lou et~al.(2023)Lou, Lu, Dai, Jia, Lin, Han, Sun, and Wu}]{USM}
Jie Lou, Yaojie Lu, Dai Dai, Wei Jia, Hongyu Lin, Xianpei Han, Le~Sun, and Hua
  Wu. 2023.
\newblock \href {http://arxiv.org/abs/2301.03282} {Universal information
  extraction as unified semantic matching}.

\bibitem[{Lu et~al.(2021)Lu, Lin, Xu, Han, Tang, Li, Sun, Liao, and
  Chen}]{Lu2021Text2EventCS}
Yaojie Lu, Hongyu Lin, Jin Xu, Xianpei Han, Jialong Tang, Annan Li, Le~Sun,
  M.~Liao, and Shaoyi Chen. 2021.
\newblock Text2event: Controllable sequence-to-structure generation for
  end-to-end event extraction.
\newblock \emph{ArXiv}, abs/2106.09232.

\bibitem[{Lu et~al.(2022)Lu, Liu, Dai, Xiao, Lin, Han, Sun, and Wu}]{UIE}
Yaojie Lu, Qing Liu, Dai Dai, Xinyan Xiao, Hongyu Lin, Xianpei Han, Le~Sun, and
  Hua Wu. 2022.
\newblock \href {http://arxiv.org/abs/2203.12277} {Unified structure generation
  for universal information extraction}.

\bibitem[{Luan et~al.(2018)Luan, He, Ostendorf, and Hajishirzi}]{SciERCDATASET}
Yi~Luan, Luheng He, Mari Ostendorf, and Hannaneh Hajishirzi. 2018.
\newblock \href {http://arxiv.org/abs/1808.09602} {Multi-task identification of
  entities, relations, and coreference for scientific knowledge graph
  construction}.

\bibitem[{Mishra et~al.(2022)Mishra, Khashabi, Baral, and
  Hajishirzi}]{mishra-etal-2022-cross}
Swaroop Mishra, Daniel Khashabi, Chitta Baral, and Hannaneh Hajishirzi. 2022.
\newblock \href {https://doi.org/10.18653/v1/2022.acl-long.244} {Cross-task
  generalization via natural language crowdsourcing instructions}.
\newblock In \emph{Proceedings of the 60th Annual Meeting of the Association
  for Computational Linguistics (Volume 1: Long Papers)}, pages 3470--3487,
  Dublin, Ireland. Association for Computational Linguistics.

\bibitem[{OpenAI(2023)}]{GPT4}
OpenAI. 2023.
\newblock Gpt-4 technical report.
\newblock \emph{ArXiv}, abs/2303.08774.

\bibitem[{openbiocorpora(2015)}]{AnatEM}
openbiocorpora. 2015.
\newblock \href {github.com/openbiocorpora/anatem} {openbiocorpora anatem}.

\bibitem[{Ouyang et~al.(2022)Ouyang, Wu, Jiang, Almeida, Wainwright, Mishkin,
  Zhang, Agarwal, Slama, Ray, Schulman, Hilton, Kelton, Miller, Simens, Askell,
  Welinder, Christiano, Leike, and Lowe}]{InstructGPT}
Long Ouyang, Jeff Wu, Xu~Jiang, Diogo Almeida, Carroll~L. Wainwright, Pamela
  Mishkin, Chong Zhang, Sandhini Agarwal, Katarina Slama, Alex Ray, John
  Schulman, Jacob Hilton, Fraser Kelton, Luke~E. Miller, Maddie Simens, Amanda
  Askell, Peter Welinder, Paul~Francis Christiano, Jan Leike, and Ryan~J. Lowe.
  2022.
\newblock Training language models to follow instructions with human feedback.
\newblock \emph{ArXiv}, abs/2203.02155.

\bibitem[{Pan et~al.(2017{\natexlab{a}})Pan, Zhang, May, Nothman, Knight, and
  Ji}]{wikiannDataset}
Xiaoman Pan, Boliang Zhang, Jonathan May, Joel Nothman, Kevin Knight, and Heng
  Ji. 2017{\natexlab{a}}.
\newblock \href {https://doi.org/10.18653/v1/P17-1178} {Cross-lingual name
  tagging and linking for 282 languages}.
\newblock In \emph{Proceedings of the 55th Annual Meeting of the Association
  for Computational Linguistics (Volume 1: Long Papers)}, pages 1946--1958,
  Vancouver, Canada. Association for Computational Linguistics.

\bibitem[{Pan et~al.(2017{\natexlab{b}})Pan, Zhang, May, Nothman, Knight, and
  Ji}]{wikiannSOTA}
Xiaoman Pan, Boliang Zhang, Jonathan May, Joel Nothman, Kevin Knight, and Heng
  Ji. 2017{\natexlab{b}}.
\newblock \href {https://doi.org/10.18653/v1/P17-1178} {Cross-lingual name
  tagging and linking for 282 languages}.
\newblock In \emph{Proceedings of the 55th Annual Meeting of the Association
  for Computational Linguistics (Volume 1: Long Papers)}, pages 1946--1958,
  Vancouver, Canada. Association for Computational Linguistics.

\bibitem[{Parmar et~al.(2022)Parmar, Mishra, Purohit, Luo, Mohammad, and
  Baral}]{parmar-etal-2022-boxbart}
Mihir Parmar, Swaroop Mishra, Mirali Purohit, Man Luo, Murad Mohammad, and
  Chitta Baral. 2022.
\newblock \href {https://doi.org/10.18653/v1/2022.findings-naacl.10}
  {In-{B}o{XBART}: Get instructions into biomedical multi-task learning}.
\newblock In \emph{Findings of the Association for Computational Linguistics:
  NAACL 2022}, pages 112--128, Seattle, United States. Association for
  Computational Linguistics.

\bibitem[{Poolsawad et~al.(2014)Poolsawad, Kambhampati, and
  Cleland}]{poolsawad2014balancing}
N~Poolsawad, C~Kambhampati, and JGF Cleland. 2014.
\newblock Balancing class for performance of classification with a clinical
  dataset.
\newblock In \emph{proceedings of the World Congress on Engineering}, volume~1,
  pages 1--6.

\bibitem[{Riedel et~al.(2010)Riedel, Yao, and McCallum}]{Riedel2010ModelingRA}
Sebastian Riedel, Limin Yao, and Andrew McCallum. 2010.
\newblock Modeling relations and their mentions without labeled text.
\newblock In \emph{ECML/PKDD}.

\bibitem[{Roth and tau Yih(2004)}]{Roth2004ALP}
Dan Roth and Wen tau Yih. 2004.
\newblock A linear programming formulation for global inference in natural
  language tasks.
\newblock In \emph{Conference on Computational Natural Language Learning}.

\bibitem[{Sang and Meulder(2003)}]{CoNLL03Dataset}
Erik F. Tjong~Kim Sang and Fien~De Meulder. 2003.
\newblock \href {http://arxiv.org/abs/cs/0306050} {Introduction to the
  conll-2003 shared task: Language-independent named entity recognition}.

\bibitem[{Schweter and Akbik(2020)}]{FindVehicleSOTA}
Stefan Schweter and Alan Akbik. 2020.
\newblock \href {http://arxiv.org/abs/2011.06993} {{FLERT:} document-level
  features for named entity recognition}.
\newblock \emph{CoRR}, abs/2011.06993.

\bibitem[{Sun et~al.(2022)Sun, Li, Pergola, Wallace, John, Greene, Kim, and
  He}]{Sun2022PHEEAD}
Zhao-Li Sun, Jiazheng Li, Gabriele Pergola, Byron~C. Wallace, Bino John, Nigel
  Greene, Joseph Kim, and Yulan He. 2022.
\newblock Phee: A dataset for pharmacovigilance event extraction from text.
\newblock \emph{ArXiv}, abs/2210.12560.

\bibitem[{Takanobu et~al.(2018)Takanobu, Zhang, Liu, and
  Huang}]{Takanobu2018AHF}
Ryuichi Takanobu, Tianyang Zhang, Jiexi Liu, and Minlie Huang. 2018.
\newblock A hierarchical framework for relation extraction with reinforcement
  learning.
\newblock In \emph{AAAI Conference on Artificial Intelligence}.

\bibitem[{Tang et~al.(2022)Tang, Zhang, He, Xu, Chao, and Xu}]{MITMovieSOTA}
Minghao Tang, Peng Zhang, Yongquan He, Yongxiu Xu, Chengpeng Chao, and Hongbo
  Xu. 2022.
\newblock \href {https://aclanthology.org/2022.coling-1.188} {{D}o{SEA}: A
  domain-specific entity-aware framework for cross-domain named entity
  recogition}.
\newblock In \emph{Proceedings of the 29th International Conference on
  Computational Linguistics}, pages 2147--2156, Gyeongju, Republic of Korea.
  International Committee on Computational Linguistics.

\bibitem[{Tedeschi et~al.(2021)Tedeschi, Maiorca, Campolungo, Cecconi, and
  Navigli}]{wikineuralDATASET}
Simone Tedeschi, Valentino Maiorca, Niccol{\`o} Campolungo, Francesco Cecconi,
  and Roberto Navigli. 2021.
\newblock \href {https://doi.org/10.18653/v1/2021.findings-emnlp.215}
  {{W}iki{NE}u{R}al: {C}ombined neural and knowledge-based silver data creation
  for multilingual {NER}}.
\newblock In \emph{Findings of the Association for Computational Linguistics:
  EMNLP 2021}, pages 2521--2533, Punta Cana, Dominican Republic. Association
  for Computational Linguistics.

\bibitem[{Tedeschi and Navigli(2022{\natexlab{a}})}]{multiNERDDATASET}
Simone Tedeschi and Roberto Navigli. 2022{\natexlab{a}}.
\newblock \href {https://doi.org/10.18653/v1/2022.findings-naacl.60}
  {{M}ulti{NERD}: A multilingual, multi-genre and fine-grained dataset for
  named entity recognition (and disambiguation)}.
\newblock In \emph{Findings of the Association for Computational Linguistics:
  NAACL 2022}, pages 801--812, Seattle, United States. Association for
  Computational Linguistics.

\bibitem[{Tedeschi and Navigli(2022{\natexlab{b}})}]{multiNERDSOTA}
Simone Tedeschi and Roberto Navigli. 2022{\natexlab{b}}.
\newblock \href {https://doi.org/10.18653/v1/2022.findings-naacl.60}
  {{M}ulti{NERD}: A multilingual, multi-genre and fine-grained dataset for
  named entity recognition (and disambiguation)}.
\newblock In \emph{Findings of the Association for Computational Linguistics:
  NAACL 2022}, pages 801--812, Seattle, United States. Association for
  Computational Linguistics.

\bibitem[{Theodoropoulos and Moens(2023)}]{ADEcorpusSOTA}
Christos Theodoropoulos and Marie-Francine Moens. 2023.
\newblock \href {http://arxiv.org/abs/2303.15100} {An information extraction
  study: Take in mind the tokenization!}

\bibitem[{Ushio and Camacho-Collados(2021)}]{MITRestaurantSOTA}
Asahi Ushio and Jose Camacho-Collados. 2021.
\newblock \href {https://doi.org/10.18653/v1/2021.eacl-demos.7} {T-{NER}: An
  all-round python library for transformer-based named entity recognition}.
\newblock In \emph{Proceedings of the 16th Conference of the European Chapter
  of the Association for Computational Linguistics: System Demonstrations}.
  Association for Computational Linguistics.

\bibitem[{Ushio et~al.(2022{\natexlab{a}})Ushio, Neves, Silva, Barbieri, and
  Camacho-Collados}]{tweetNER7DATASET}
Asahi Ushio, Leonardo Neves, Vitor Silva, Francesco Barbieri, and Jose
  Camacho-Collados. 2022{\natexlab{a}}.
\newblock \href {http://arxiv.org/abs/2210.03797} {Named entity recognition in
  twitter: A dataset and analysis on short-term temporal shifts}.

\bibitem[{Ushio et~al.(2022{\natexlab{b}})Ushio, Neves, Silva, Barbieri, and
  Camacho-Collados}]{tweetNER7SOTA}
Asahi Ushio, Leonardo Neves, Vitor Silva, Francesco Barbieri, and Jose
  Camacho-Collados. 2022{\natexlab{b}}.
\newblock \href {http://arxiv.org/abs/2210.03797} {Named entity recognition in
  twitter: A dataset and analysis on short-term temporal shifts}.

\bibitem[{Walker and Consortium(2005)}]{ACE2005DATASET}
C.~Walker and Linguistic~Data Consortium. 2005.
\newblock \href {https://books.google.com/books?id=SbjjuQEACAAJ} {\emph{ACE
  2005 Multilingual Training Corpus}}.
\newblock LDC corpora. Linguistic Data Consortium.

\bibitem[{Wang et~al.(2023)Wang, Liu, Chen, Hong, Tang, and
  Song}]{GENIANERSOTA}
Chenguang Wang, Xiao Liu, Zui Chen, Haoyun Hong, Jie Tang, and Dawn Song. 2023.
\newblock \href {http://arxiv.org/abs/2205.10475} {Deepstruct: Pretraining of
  language models for structure prediction}.

\bibitem[{Wang et~al.(2022{\natexlab{a}})Wang, Li, Yan, Yan, Wang, Wu, and
  Xu}]{Wang2022InstructionNERAM}
Liwen Wang, Rumei Li, Yang Yan, Yuanmeng Yan, Sirui Wang, Wei~Yu Wu, and Weiran
  Xu. 2022{\natexlab{a}}.
\newblock Instructionner: A multi-task instruction-based generative framework
  for few-shot ner.
\newblock \emph{ArXiv}, abs/2203.03903.

\bibitem[{Wang et~al.(2022{\natexlab{b}})Wang, Dou, Xiong, Zou, Zhang, Gui,
  Qiao, Cheng, and Huang}]{wang-etal-2022-miner}
Xiao Wang, Shihan Dou, Limao Xiong, Yicheng Zou, Qi~Zhang, Tao Gui, Liang Qiao,
  Zhanzhan Cheng, and Xuanjing Huang. 2022{\natexlab{b}}.
\newblock \href {https://doi.org/10.18653/v1/2022.acl-long.383} {{MINER}:
  Improving out-of-vocabulary named entity recognition from an information
  theoretic perspective}.
\newblock In \emph{Proceedings of the 60th Annual Meeting of the Association
  for Computational Linguistics (Volume 1: Long Papers)}, pages 5590--5600,
  Dublin, Ireland. Association for Computational Linguistics.

\bibitem[{Wang et~al.(2021{\natexlab{a}})Wang, Jiang, Bach, Wang, Huang, Huang,
  and Tu}]{CoNLL2003SOTA}
Xinyu Wang, Yong Jiang, Nguyen Bach, Tao Wang, Zhongqiang Huang, Fei Huang, and
  Kewei Tu. 2021{\natexlab{a}}.
\newblock \href {https://doi.org/10.18653/v1/2021.acl-long.206} {Automated
  concatenation of embeddings for structured prediction}.
\newblock In \emph{Proceedings of the 59th Annual Meeting of the Association
  for Computational Linguistics and the 11th International Joint Conference on
  Natural Language Processing (Volume 1: Long Papers)}, pages 2643--2660,
  Online. Association for Computational Linguistics.

\bibitem[{Wang et~al.(2021{\natexlab{b}})Wang, Sun, Wu, Zhou, Li, and
  Yan}]{broad_twitter_corpusSOTA}
Yijun Wang, Changzhi Sun, Yuanbin Wu, Hao Zhou, Lei Li, and Junchi Yan.
  2021{\natexlab{b}}.
\newblock \href {https://doi.org/10.18653/v1/2021.acl-long.19} {{U}ni{RE}: A
  unified label space for entity relation extraction}.
\newblock In \emph{Proceedings of the 59th Annual Meeting of the Association
  for Computational Linguistics and the 11th International Joint Conference on
  Natural Language Processing (Volume 1: Long Papers)}, pages 220--231, Online.
  Association for Computational Linguistics.

\bibitem[{Wang et~al.(2022{\natexlab{c}})Wang, Mishra, Alipoormolabashi, Kordi,
  Mirzaei, Naik, Ashok, Dhanasekaran, Arunkumar, Stap, Pathak, Karamanolakis,
  Lai, Purohit, Mondal, Anderson, Kuznia, Doshi, Pal, Patel, Moradshahi,
  Parmar, Purohit, Varshney, Kaza, Verma, Puri, Karia, Doshi, Sampat, Mishra,
  Reddy~A, Patro, Dixit, and Shen}]{wang-etal-2022-super}
Yizhong Wang, Swaroop Mishra, Pegah Alipoormolabashi, Yeganeh Kordi, Amirreza
  Mirzaei, Atharva Naik, Arjun Ashok, Arut~Selvan Dhanasekaran, Anjana
  Arunkumar, David Stap, Eshaan Pathak, Giannis Karamanolakis, Haizhi Lai,
  Ishan Purohit, Ishani Mondal, Jacob Anderson, Kirby Kuznia, Krima Doshi,
  Kuntal~Kumar Pal, Maitreya Patel, Mehrad Moradshahi, Mihir Parmar, Mirali
  Purohit, Neeraj Varshney, Phani~Rohitha Kaza, Pulkit Verma, Ravsehaj~Singh
  Puri, Rushang Karia, Savan Doshi, Shailaja~Keyur Sampat, Siddhartha Mishra,
  Sujan Reddy~A, Sumanta Patro, Tanay Dixit, and Xudong Shen.
  2022{\natexlab{c}}.
\newblock \href {https://aclanthology.org/2022.emnlp-main.340}
  {Super-{N}atural{I}nstructions: Generalization via declarative instructions
  on 1600+ {NLP} tasks}.
\newblock In \emph{Proceedings of the 2022 Conference on Empirical Methods in
  Natural Language Processing}, pages 5085--5109, Abu Dhabi, United Arab
  Emirates. Association for Computational Linguistics.

\bibitem[{Xie et~al.(2021)Xie, Liang, Liu, Huang, Huang, and
  Xiao}]{NYT11HRLSOTA}
Chenhao Xie, Jiaqing Liang, Jingping Liu, Chengsong Huang, Wenhao Huang, and
  Yanghua Xiao. 2021.
\newblock \href {http://arxiv.org/abs/2105.10158} {Revisiting the negative data
  of distantly supervised relation extraction}.
\newblock \emph{CoRR}, abs/2105.10158.

\bibitem[{Yan et~al.(2021)Yan, Dai, Ji, Qiu, and Zhang}]{yan-etal-2021-unified}
Hang Yan, Junqi Dai, Tuo Ji, Xipeng Qiu, and Zheng Zhang. 2021.
\newblock \href {https://doi.org/10.18653/v1/2021.acl-long.188} {A unified
  generative framework for aspect-based sentiment analysis}.
\newblock In \emph{Proceedings of the 59th Annual Meeting of the Association
  for Computational Linguistics and the 11th International Joint Conference on
  Natural Language Processing (Volume 1: Long Papers)}, pages 2416--2429,
  Online. Association for Computational Linguistics.

\bibitem[{Ye et~al.(2021)Ye, Lin, and Sun}]{SciERCSOTA}
Deming Ye, Yankai Lin, and Maosong Sun. 2021.
\newblock \href {http://arxiv.org/abs/2109.06067} {Pack together: Entity and
  relation extraction with levitated marker}.
\newblock \emph{CoRR}, abs/2109.06067.

\bibitem[{Ye et~al.(2023)Ye, Chen, Xu, Zu, Shao, Liu, Cui, Zhou, Gong, Shen
  et~al.}]{ye2023comprehensive}
Junjie Ye, Xuanting Chen, Nuo Xu, Can Zu, Zekai Shao, Shichun Liu, Yuhan Cui,
  Zeyang Zhou, Chao Gong, Yang Shen, et~al. 2023.
\newblock A comprehensive capability analysis of gpt-3 and gpt-3.5 series
  models.
\newblock \emph{arXiv preprint arXiv:2303.10420}.

\bibitem[{Zhang and Wang(2015)}]{kbp37DATASET}
Dongxu Zhang and Dong Wang. 2015.
\newblock \href {http://arxiv.org/abs/1508.01006} {Relation classification via
  recurrent neural network}.

\bibitem[{Zhang et~al.(2023)Zhang, Cheng, Gao, and Poon}]{bc5cdrSOTA}
Sheng Zhang, Hao Cheng, Jianfeng Gao, and Hoifung Poon. 2023.
\newblock \href {http://arxiv.org/abs/2208.14565} {Optimizing bi-encoder for
  named entity recognition via contrastive learning}.

\bibitem[{Zhong and Chen(2020)}]{ACE0405SOTA}
Zexuan Zhong and Danqi Chen. 2020.
\newblock \href {http://arxiv.org/abs/2010.12812} {A frustratingly easy
  approach for joint entity and relation extraction}.
\newblock \emph{CoRR}, abs/2010.12812.

\end{thebibliography}
\bibliographystyle{acl_natbib}

\section{Appendix}
\label{sec:appendix}
\subsection{Data Details}
\label{data details}
IE INSTRUCTIONS collects 32 publicly available datasets covering three IE tasks: NER, RE, and EE.
For NER(named entity extraction) task, the 21 used datasets includes ACE2004, ACE2005\cite{ACE2005DATASET}, broad\_twitter\_corpus\cite{broad_twitter_corpusDATASET}, CoNLL2003\cite{CoNLL03Dataset}, multiNERD\cite{multiNERDDATASET}, Ontonotes\cite{OntoNotesDataset}, polyglot-NER\cite{polyglot-NERDATASET}, tweetNER7\cite{tweetNER7DATASET}, wikiann\cite{wikiannDataset}, wikineural\cite{wikineuralDATASET}, AnatEM\cite{AnatEM}, bc2gm\cite{Kocaman2020BiomedicalNE}, bc4chemd\cite{bc4chemdDATASET}, bc5cdr\cite{Li2016BioCreativeVC}, FabNER\cite{Kumar2021FabNERIE}, FindVehicle\cite{FindVehicle}, GENIA\cite{GENIANERDATASET}, HarveyNER\cite{HarveyNERDATASET}, MIT Movie Review\cite{MITReviewDataset}, MIT Restaurant Review\cite{MITReviewDataset} and ncbi-disease\cite{ncbi-diseaseDATASET}. For RE(relation extraction) task, we use 10 datasets including ADE corpus\cite{ADEcorpusDATASET}, CoNLL2004\cite{Roth2004ALP}, GIDS\cite{Jat2018ImprovingDS}, kbp37\cite{kbp37DATASET}, NYT\cite{Riedel2010ModelingRA}, NYT11 HRL\cite{Takanobu2018AHF}, SciERC\cite{SciERCDATASET}, semeval RE\cite{Hendrickx2010SemEval2010T8}, FewRel\cite{FewRelDATASET} and Wiki-ZSL\cite{Wiki-ZSLDATASET}. For task EE(event extraction), ACE2005\cite{ACE2005DATASET}, CASIE\cite{Lu2021Text2EventCS}, GENIA\cite{Kim2003GENIAC} and PHEE\cite{Sun2022PHEEAD} are used. 

For the data set with only training set as the original data, we divided it into training set, validation set and test set according to the ratio of 8:1:1. For the data set with only training set and validation set as the original data, we randomly select half of the data in the validation set as the test set and the other half as the new validation set. For other datasets, we adopt the official split.

Tabel \ref{dataset-details} shows detailed datasets statistics. NER refers to Named Entity Recognition task, RE refers to Relation Extraction task, and EE refers to Event Extraction task. |Labels| indicates the number of labels, and $\#$ is the number of sentences in the specific subset.
 For the |Labels| of event extraction, the number outside the parenthesis indicates the number of event types and the number inside the parenthesis indicates the number of argument types.
\begin{table*}[htbp]
    \centering
    \begin{tabular}{c|c|c|ccc}
    \toprule  
    Task & Dataset & |labels| & \#Train & \#Val & \#Test \\
    \midrule
    \multirow{26}*{NER} & ACE2004 & 7 & 6202 & 745 & 812 \\
    ~ & ACE2005 & 7 & 7299 & 971 & 1060 \\
    ~ & broad\_twitter\_corpus & 3 & 5334 & 2000 & 2001 \\
    ~ & CoNLL2003 & 4 & 14041 & 3250 & 3453 \\
    ~ & multiNERD & 16 & 134144 & 10000 & 10000 \\
    ~ & Ontonotes & 18 & 59924 & 8528 & 8262\\
    ~ & polyglot-NER & 3 & 393982 & 10000 & 10000 \\
    ~ & tweetNER7 & 7 & 7111 & 886 & 576 \\
    ~ & wikiann & 3 & 20000 & 10000 & 10000 \\
    ~ & wikineural & 3 & 92720 & 11590 & 11597 \\
    ~ & AnatEM & 1 & 5861 & 2118 & 3830 \\
    ~ & bc2gm & 1 & 12500 & 2500 & 5000 \\
    ~ & bc4chemd & 1 & 30682 & 30639 & 26364 \\
    ~ & bc5cd & 2 & 4560 & 4581 & 4797 \\
    ~ & CrossNER\_AI & 14 & 100 & 350 & 431 \\
    ~ & CrossNER\_literature & 12 & 100 & 400 & 416 \\
    ~ & CrossNER\_music & 13 & 100 & 380 & 465 \\
    ~ & CrossNER\_politics & 9 & 199 & 540 & 650 \\
    ~ & CrossNER\_science & 17 & 200 & 450 & 543 \\
    ~ & FabNER & 12 & 9435 & 2182 & 2064 \\
    ~ & FindVehicle & 21 & 21565 & 20777 & 20777 \\
    ~ & GENIA & 5 & 15023 & 1669 & 1854 \\
    ~ & HarveyNER & 4 & 3967 & 1301 & 1303 \\
    ~ & MIT Movie Review & 12 & 9774 & 2442 & 2442 \\
    ~ & MIT Restaurant Review & 8 & 7659 & 1520 & 1520 \\
    ~ & ncbi-disease & 1 & 5432 & 923 & 940 \\
    \midrule
    \multirow{10}*{RE} & ADE corpus & 1 & 3417 & 427 & 428 \\
    ~ & CoNLL2004 & 5 & 922 & 231 & 288 \\
    ~ & GIDS & 4 & 8526 & 1417 & 4307 \\
    ~ & kbp37 & 18 & 15917 & 1724 & 3405 \\
    ~ & NYT & 24 & 56196 & 5000 & 5000 \\
    ~ & NYT11 HRL & 12 & 62648 & 149 & 369 \\
    ~ & SciERC & 7 & 1366 & 187 & 397 \\
    ~ & semeval RE & 10 & 6507 & 1493 & 2717 \\
    \midrule
    \multirow{4}*{EE} & ACE2005 & 33(22) & 3342 & 327 & 293 \\
    ~ & CASIE & 5(26) & 3751 & 788 & 1500 \\
    ~ & GENIA & 5(0) & 15023 & 1669 & 1854 \\
    ~ & PHEE & 2(16) & 2898 & 961 & 968 \\
    \bottomrule
    \end{tabular}
    \caption{\label{dataset-details}
Detailed datasets statistics.}
\end{table*}

\subsection{Instruction Details}
Table \ref{prompts_details} shows prompts for different tasks. NER refers to the named entity recognition task, the object of which is the entity in the output sentence and its corresponding entity type. RE refers to the relation extraction task, the object of which is to extract the relation triplet in the sentence, including the relation name, the head entity and the tail entity. EE refers to the event extraction task. The task objective is to extract the event types, trigger word and arguments in the sentence. ES refers to entity span, the task target is given sentence and entity category options, and output entities that conform to the entity category, but there is no need to output the entity type of each entity; ET refers to entity type identification. The task target is a given sentence, which contains entity and entity category options, and outputs the entity category corresponding to each entity. EP refers to entity pair identification (entity pair). The task target is given sentence and relation category options, and output entity pairs that conform to relation category, but do not need to output its relation category; EPR refers to entity pair relationship identification. The task target is a given sentence, which contains entity pair and relationship category options, and outputs the corresponding relationship category for each entity pair. ES and ET are auxiliary tasks of NER, EP and EPR are auxiliary tasks of RE, and EEA and EET are auxiliary tasks of EE.

\begin{table*}[htbp]
    \centering
    \begin{tabular}{m{0.1\linewidth}m{0.85\linewidth}}
    \toprule    
        \textbf{Task} & \textbf{{Prompts}} \\ \midrule
        \multirow{7}*{NER} & Please list all entity words in the text that fit the category. Output format is "type1: word1; type2: word2". \\ \\
        ~ & Please find all the entity words associated with the category in the given text. Output format is "type1: word1; type2: word2". \\ \\
        ~ & Please tell me all the entity words in the text that belong to a given category. Output format is "type1: word1; type2: word2". \\ 
        
        \midrule
        \multirow{8}*{RE} & Given a phrase that describes the relationship between two words, extract the words and the lexical relationship between them. The output format should be "relation1: word1, word2; relation2: word3, word4". \\ \\
        ~ & Find the phrases in the following sentences that have a given relationship. The output format is "relation1: word1, word2; relation2: word3, word4". \\ \\
        ~ & Given a sentence, please extract the subject and object containing a certain relation in the sentence according to the following relation types, in the format of "relation1: word1, word2; relation2: word3, word4". \\ 
        
        \midrule
        \multirow{3}*{EE} & Locate the role in the text that participated in the event based on the event type and return it in the event list. \\ \\
        ~ & Extract the event information in the text and return them in the event list. \\ 
        
        \midrule
        \multirow{1}*{ES} & Please list all entity words in the text that fit the category. Output format is word1, word2. \\ 
        
        \midrule
        \multirow{1}*{ET} & Given options, please tell me the categories of all the listed entity words.Output format is "type1: word1; type2: word2". \\ 
        
        \midrule
        \multirow{1}*{EP} & Please list all entity pairs containing a certain relationship in the given options.Output format is "word1, word2; word3, word4". \\ 
        
        \midrule
        \multirow{1}*{EPR} & Given options, please tell me the relationships of all the listed entity pairs.Output format is "relation1: word1, word2; relation2: word3, word4". \\ 
        
        \midrule
        \multirow{1}*{EEA} & Given event type and trigger, please tell me the arguments of all the listed option. Output format is "name: role". \\ 
        
        \midrule
        \multirow{1}*{EET} & Please tell me event type and its trigger word from given type options. Output format is "event type: trigger". \\ \midrule
    \end{tabular}
    \caption{\label{prompts_details}
    Instructions for different tasks.
    }
\end{table*}

\begin{table*}[htbp]
    \centering
    \resizebox{\textwidth}{!}{
    \begin{tabular}{c|c|cc}
    \toprule  
    Dataset & Metric & \multicolumn{2}{c}{Task-specific SOTA Methods} \\
    \midrule
    ACE2004 & Entity F1 & \citet{ACE0405SOTA} & 90.3 \\
    ACE2005-Ent & Entity F1 & \citet{ACE0405SOTA} & 90.9 \\
    AnatEM & Entity F1 & \citet{AnatEMSOTA} & 91.65 \\
    bc2gm & Entity F1 & \citet{bc2gmSOTA} & 88.75 \\
    bc4chemd & Entity F1 & \citet{bc4chemdSOTA} & 94.39 \\
    bc5cdr & Entity F1 & \citet{bc5cdrSOTA} & 91.9 \\
    broad\_twitter\_corpus & Entity F1 & \citet{broad_twitter_corpusSOTA} & 74.70 \\
    CoNLL2003 & Entity F1 & \citet{CoNLL2003SOTA} & 94.60\\
    FabNER & Entity F1 & \citet{FabNERSOTA} & 88 \\
    FindVehicle & Entity F1 & \citet{FindVehicleSOTA} & 80.9 \\
    GENIA-Ent & Entity F1 & \citet{GENIANERSOTA} & 80.80 \\
    HarveyNER & Entity F1 & \citet{HarveyNERSOTA} & 68.97 \\
    MIT Movie Review & Entity F1 & \citet{MITMovieSOTA} & 87.31 \\
    MIT Restaurant Review & Entity F1 & \citet{MITRestaurantSOTA} & 79.6 \\
    multiNERD & Entity F1 & \citet{multiNERDSOTA} & 85.0 \\
    ncbi-disease & Entity F1 & \citet{ncbiSOTA} & 90.48 \\
    Ontonotes & Entity F1 & \citet{OntonotesSOTA} & 92.07 \\
    polyglot-NER & Entity F1 & - \\
    tweetNER7 & Entity F1 & \citet{tweetNER7SOTA} & 66 \\
    wikiann & Entity F1 & \citet{wikiannSOTA} & 91.8 \\
    wikineural & Entity F1 & -  \\
    ADE corpus & Relation Strict F1 & \citet{ADEcorpusSOTA} & 83.9 \\
    CoNLL2004 & Relation Strict F1 & \citet{CoNLL2004SOTA} & 76.65 \\
    GIDS & Relation Strict F1 & - \\
    kbp37 & Relation Strict F1 & - \\
    NYT & Relation Strict F1 & - \\
    NYT11 HRL & Relation Strict F1 & \citet{NYT11HRLSOTA} & 55.47 \\
    SciERC & Relation Strict F1 & \citet{SciERCSOTA} & 38.40 \\ 
    semeval RE & Relation Strict F1 & \
    \citet{semevalRESOTA} & 76.00 \\
    ACE2005 & Event Trigger F1 & - & -  \\
    ACE2005 & Event Argument F1 & - & - \\
    CASIE & Event Trigger F1 & - & - \\
    CASIE & Event Argument F1 & - & - \\
    GENIA-Evt & Event Trigger F1 & - & 63.96 \\
    GENIA-Evt & Event Argument F1 & - & - \\
    PHEE & Event Trigger F1 & - & - \\
    PHEE & Event Argument F1 & - & - \\
    \bottomrule
    \end{tabular}
}
    \caption{
Overall results of InstructUIE on different datasets. InstructUIE perform better or comparable than Bert on popular NER datasets like ACE2005, CoNLL2003, Ontonotes, and tweetNER7. In the RE task, InstructUIE achieved results comparable to the baseline on most datasets. In the EE task, our model outperformed USM, UIE or SOTA on some datasets.
}
\label{supervised-result-withSOTA}
\end{table*}

\end{document}